\documentclass[sigconf,nonacm]{acmart}

\usepackage{times}
\usepackage{latexsym}
\usepackage{multirow}
\usepackage{booktabs} 
\usepackage{url}
\usepackage{amsmath,amssymb,amsfonts}
\usepackage{graphicx}
\usepackage{graphics}
\usepackage{enumitem} 
\usepackage{multirow}
\usepackage[normalem]{ulem}
\useunder{\uline}{\ul}{}
\usepackage{algorithm,algpseudocode}
\usepackage{booktabs} 

\begin{document}
\title{Model Compression with Two-stage Multi-teacher Knowledge Distillation for Web Question Answering System}
\titlenote{This paper has been accepted by WSDM 2020.}


\author{Ze Yang$^\dag$}
 \email{yaze@microsoft.com}
\affiliation{
	\department[0]{STCA NLP Group, Microsoft} 
 }
 \author{Linjun Shou}
 	\authornote{These authors contributed equally.}
 \email{lisho@microsoft.com}
\affiliation{
	\department[0]{STCA NLP Group, Microsoft} 
 }
 \author{Ming Gong}
 \email{migon@microsoft.com}
\affiliation{
	\department[0]{STCA NLP Group, Microsoft} 
 }
 \author{Wutao Lin}
 \email{wutlin@microsoft.com}
\affiliation{
	\department[0]{STCA NLP Group, Microsoft} 
 }
 \author{Daxin Jiang}
 \email{djiang@microsoft.com}
\affiliation{
	\department[0]{STCA NLP Group, Microsoft} 
 }

\renewcommand{\shortauthors}{}

\begin{abstract}
    Deep pre-training and fine-tuning models (such as BERT and OpenAI GPT) have demonstrated excellent results in question answering areas. However, due to the sheer amount of model parameters, the inference speed of these models is very slow. How to apply these complex models to real business scenarios becomes a challenging but practical problem. Previous model compression methods usually suffer from information loss during the model compression procedure, leading to inferior models compared with the original one. To tackle this challenge, we propose a Two-stage Multi-teacher Knowledge Distillation (TMKD for short) method for web Question Answering system. We first develop a general Q\&A distillation task for student model pre-training, and further fine-tune this pre-trained student model with multi-teacher knowledge distillation on downstream tasks (like Web Q\&A task, MNLI, SNLI, RTE tasks from GLUE), which effectively reduces the overfitting bias in individual teacher models, and transfers more general knowledge to the student model. The experiment results show that our method can significantly outperform the baseline methods and even achieve comparable results with the original teacher models, along with substantial speedup of model inference.






\end{abstract}

%
%


\keywords{model compression, two-stage, multi-teacher, knowledge distillation, distillation pre-training}

\maketitle

\section{Introduction}
Question Answering relevance, which aims to rank the text passages to natural language questions issued by users, is a critical task in Question Answering (Q\&A) system \cite{cimiano2014ontology}. In recent years, almost all commercial web search engines provide Question Answering service, in addition to the traditional web documents links. Table~\ref{t:example} shows an example for Question Answering from a commercial search engine. Compared with the ``ten-blue-links'', Q\&A is a more natural interface, and thousands of millions of users enjoy the efficiency of directly accessing the information for their questions.

\begin{table}[htbp]
    \renewcommand{\arraystretch}{1.5}
    \centering
    \caption{\label{t:example}An example of Q\&A relevance task.}
    \begin{tabular}{lp{6cm}p{7cm}}
    \hline
    \textbf{Question}: &\emph{What can I do when I have headache?}\\
    \hline
    \textbf{Passage}: &\emph{Drinking warm water mixed with juice squeezed from one-half of a lemon will reduce the intensity of a headache. This particular remedy is beneficial for headaches caused by gas in the stomach. Another option is to apply lemon crusts, pounded into a paste, on your forehead to immediately relieve pain...} \\ \hline
    \textbf{Label}: &\emph{Relevant} \\
    \hline
    \end{tabular}
\end{table}


In recent years, deep pre-training approaches~\cite{radford2018improving, devlin2018bert} have brought big break-through in NLP tasks. They also show very promising results for the particular task of Q\&A relevance. However, due to the huge parameter size of these models (For example, GPT/BERT\textsubscript{base} has 110M parameters, and BERT\textsubscript{large} has 340M.), both model training and inference become very time-consuming. Although several works have studied the optimization of model training ~\cite{abs-1904-00962}, there is little work discussing the model inference challenge of deep pre-training models like BERT/GPT models. In fact, for a web scale Q\&A system, the efficiency of model inference may be even more critical than that of model training, due to the concerns of both offline throughput and online latency (we will elaborate more in the following paragraphs). 

Table~\ref{t:example_inference} shows the inference speed of BERT models~\cite{devlin2018bert} with a 1080Ti GPU. The throughout of Q\&A pairs are 624 and 192 per second on average for BERT\textsubscript{base} and BERT\textsubscript{large}, respectively. In other words, the average latency are 1.6 and 5.2 milliseconds respectively.
\begin{table}[htbp] 
    \centering
    \caption{\label{t:example_inference}The inference speed of BERT on 1080Ti GPU.}
\begin{tabular}{@{}lrrr@{}}
\toprule
\textbf{Model}      & \textbf{Parameter} & \textbf{\begin{tabular}[c]{@{}c@{}}Samples\\ Per second\end{tabular}} & \textbf{Latency} \\ \midrule
\textbf{BERT\textsubscript{base}}  & 110M               & 624                                                                    & 1.6ms           \\
\textbf{BERT\textsubscript{large}} & 340M               & 192                                                                    & 5.2ms           \\ \bottomrule
\end{tabular}
\end{table}

In a commercial web Q\&A system, there are often two complementary pipelines for the Q\&A service. One pipeline is for popular queries that frequently appear in the search traffic. The answers are pre-computed offline in a batch mode and then served online by simple look-up. The magnitude of the number of Q\&A pairs processed is around 10 billions. The other pipeline is for tail queries that are rarely or never seen before. For such tail queries, the answers are ranked on the fly and the latency budget for online model inference is typically within 10 milliseconds. Therefore, for both offline or online pipelines, it is critical to improve model inference efficiency.

To improve model inference efficiency, we consider model compression approach. In other words, we aim to train a smaller model with fewer parameters to simulate the original large model. A popular method, called {\em knowledge distillation}~\cite{hinton2015distilling} has been widely used for model compression. The basic idea is a teacher-student framework, in which the knowledge from a complex network (teacher model) is transferred to a simple network (student model) by learning the output distribution of the teacher model as a soft target. To be more specific, when training the student model, we not only provide the human-labeled golden ground truth, but also feed the output score from the teacher model as a secondary soft label. Compared with the discrete human labels (for classification task), the continuous scores from the teacher models give more smooth and fine-grained supervision to the student model, and thus result in better model performance. We refer to this basic knowledge distillation approach as {\em 1-o-1 model}, in the sense that one teacher transfers knowledge to one student.

Although the 1-o-1 model can effectively reduce the number of parameters as well as the time for model inference, due to the information loss during the knowledge distillation, the performance of student model usually cannot reach the parity with its teacher model. This motivates us to develop the second approach, called {\em m-o-m ensemble model}. To be more specific, we first train multiple teacher models, for example, BERT (base and large)~\cite{devlin2018bert} and GPT~\cite{radford2018improving} with different hyper-parameters. Then train a separate student model for each individual teacher model. Finally, the student models trained from different teachers are ensembled to generate the ultimate result. Our experimental results showed that the m-o-m ensemble model performs better than the 1-o-1 model. The rationale is as follows. Each teacher model is trained towards a specific learning objective. Therefore, various models have different generalization ability, and they also overfit the training data in different ways. When ensemble these models, the over-fitting bias across different models can be reduced by the voting effect. That say, the ensemble models automatically ``calibrate'' the results.

When we compare the m-o-m ensemble model with the 1-o-1 model, although the former has better performance, it also consumes much larger memory to host multiple student models. This motivates us to look for a new approach, which has better performance than the 1-o-1 model and consumes less memory than the m-o-m model. One observation for the m-o-m ensemble approach is that it conducts the model ensemble too late. In fact,  once the training process for a student models has finished, the overfitting bias from the corresponding teacher model has already been transferred to the student model. The voting effect across student models can be considered as a ``late calibration'' process. On the other hand, if we feed the scores from multiple teachers to a single student model during the training stage, that model is receiving guidance from various teachers simultaneously. Therefore, the overfitting bias can be addressed by ``early calibration''. Based on this observation, we develop the novel {\em m-o-1} approach, where we train a single student model by feeding the scores from multiple teachers at the same time as the supervision signals. The experimental results showed that the m-o-1 model performs better than the m-o-m model, while the memory consumption is the same with the 1-o-1 model.

The novel m-o-1 approach results in decent compressed models. However, the performance of the compressed models still has small gap with the original large model. One obvious reason is that the original large model has a large-scale pre-training stage, where it learns the language model through an unsupervised approach. We therefore explore how to simulate a pre-training stage for the compressed models, such that it can benefit from large-scale training data and learn the feature representation sufficiently. 


Our empirical study shows that the pre-training stage significantly improves the model performance. When we adopt a very large pre-training data, followed by the m-o-1 fine-tuning strategy, the compressed model can achieve comparable or even better performance than the teacher model. Another interesting finding is that although the pre-trained model is derived from Q\&A pairs, it can serve as a generic baseline for multiple tasks. As we show in the experiment part, when we fine-tune the Q\&A pre-trained model with various text matching tasks, such as those in GLUE \cite{WangSMHLB18}, it outperforms the compressed model without pre-training on each task. To the best of our knowledge, this is the first work discussing the
distillation pre-training and multiple teacher distillation for Web Q\&A.

\begin{figure*}[t!]
    \centering
    \includegraphics[scale=0.61, viewport=-20 133 965 485, clip=true]{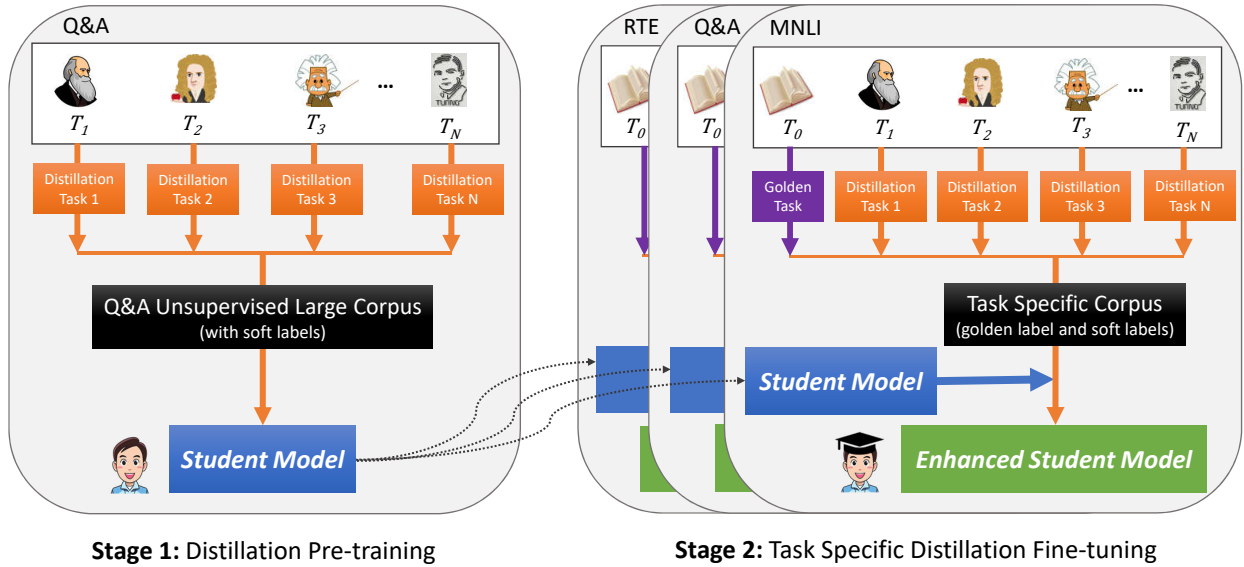}
    \vspace{-10pt}
    \caption{\label{fig:two_stage_model} The Overall Architecture of Our Two-stage Multi-teacher Distillation Model.}
\end{figure*}

In this paper, we propose a \textbf{T}wo-stage \textbf{M}ulti-teacher \textbf{K}nowledge \textbf{D}istillation (\textbf{TMKD} for short) method for model compression, and make the following major contributions.
\begin{itemize}
    \item In the first stage (i.e., the pre-training stage) of TMKD, we create a general Q\&A distillation pre-training task to leverage large-scale unlabeled question-passage pairs derived from a commercial search engine. The compressed model benefits from such large-scale data and learns feature representation sufficiently. This pre-trained Q\&A distillation model can be also applied to the model compression of various text matching tasks.
    \item In the second stage (i.e., the fine-tuning stage) of TMKD, we design a multi-teacher knowledge distillation paradigm to jointly learn from multiple teacher models on downstream tasks. The ``early calibration'' effect relieves the over-fitting bias in individual teacher models, and consequently, the compressed model can achieve comparable or even better performance with the teacher model.
    \item We conduct intensive experiments on several datasets (both open benchmark and commercial large-scale datasets) to verify the effectiveness of our proposed method. TMKD outperforms various state-of-the-art baselines and has been applied to real commercial scenarios.
\end{itemize}

The rest of the paper is organized as follows. After a summary of related work in Section 2, we describe our proposed model in details in Section 3, followed by comprehensive evaluations in Section 4 and Section 5. Finally, Section 6 concludes this paper and discuss future directions.

\section{Related Work}
In this section we briefly review two research areas related to our work: model compression and multi-task learning.

\subsection{Model Compression} 
As the parameter size of neural network model is getting larger and larger~\cite{RuderH18,PetersNIGCLZ18,devlin2018bert}, how to make it feasible to deploy and apply the models in industrial environment becomes an important problem. A natural process is to compress the model~\cite{CunDS89, hinton2015distilling, frankle2018lottery}. Low-rank approximation was a factorization method~\cite{ZhangZMHS15,jaderberg2014speeding,DentonZBLF14}, which used multiple low rank matrices to approximate the original matrix to reduce model redundancy~\cite{CunDS89,NIPS1992_647,HeZS17}. \citeauthor{hinton2015distilling} proposed a knowledge distillation method (KD for short) \shortcite{hinton2015distilling}. In their work, the output of the complex network was used as a soft target for the training of simple network. By this way, the knowledge of complex models can be transferred to simple models. Distilling complex models into simple models has been shown to improve many NLP tasks to achieve impressive performance \cite{KimR16, MouJXL0J16, KuncoroBKDS16, abs-1904-09482}. \citeauthor{abs-1802-05668} \shortcite{abs-1802-05668} proposed a quantized distillation method. In their work, they incorporated distillation loss, and expressed with respect to the teacher network, into the training process of a smaller student network whose weights were quantized to a limited set of levels. \citeauthor{PapernotAEGT17}~\cite{PapernotAEGT17} proposed a training data protected method based on knowledge distillation . In their work, an ensemble of teachers was trained on disjoint subsets of the sensitive data, and then a student model was trained on public data labeled using the ensemble of teachers. 

\subsection{Multi-task Learning} 
Multi-task learning has been widely studied in deep learning, which leverages the information among different tasks to improve the generalization performance~\cite{Yim2015Rotating, collobert2008unified, Dong2014Inferring}. \citeauthor{FaresOV18} \shortcite{FaresOV18} empirically evaluated the utility of transfer and multi-task learning on semantic interpretation of noun-noun compounds. It showed that transfer learning via parameter sharing can help a neural classification model generalize over a highly skewed distribution of relations. \citeauthor{Pentina2017Multi}~\cite{Pentina2017Multi} studied a variant of multi-task learning in which annotated data was available on some of the tasks. \citeauthor{LeePCCB15}~\cite{LeePCCB15} studied the performance of different ensemble methods under the framework of multi-task learning. 

\citeauthor{YouX0T17}~\cite{YouX0T17} presented a method to train a thin deep network by incorporating in the intermediate layers and imposing a constraint about the dissimilarity among examples. \citeauthor{WuCW19}~\cite{WuCW19} propose a multi-teacher knowledge distillation framework for compressed video action recognition to compress this model. These efforts have tried multiple teacher distillation methods in the field of computer vision, but little research has been done on the NLP deep pre-training based model. Concurrently with our work, several works also combine the multi-task learning with knowledge distillation~\cite{abs-1904-09482, LiuHCG19, ClarkLKML19}. However, they applied the knowledge distillation and multi-task learning to enhance the original model performance, instead of targeting model compression.

Our approach is also a knowledge distillation based method for model compression. Different from previous approaches, we develop a novel Q\&A distillation pre-training task leveraging large-scale unsupervised Q\&A data. Moreover, we design a multi-task paradigm in the fine-tuning stage to jointly distill the knowledge from different teacher models into a single student model. 

\section{Our Approach}
In this section, we firstly describe the overall design of our model, and then describe the proposed approach TMKD in details. Finally, we discuss the procedure of model training and prediction.

\subsection{Overview}
Figure~\ref{fig:two_stage_model} shows the architecture of TMKD. It consists of two stages: distillation pre-training and task specific distillation fine-tuning. In terms of teacher model for distillation, we take labeled data by crowd sourcing judges as one specific teacher ($T_0$) which has the ground-truth knowledge (e.g. $0$ or $1$). We also have several other teachers ($T_1$-$T_N$) trained on different pre-trained models (e.g., BERT~\cite{devlin2018bert} and GPT~\cite{radford2018improving}) or with different hyper-parameters, which provide the soft knowledge as pseudo supervision (score in $[0,1]$).  

\subsubsection{Stage 1 - Distillation Pre-training}
Deep pre-trained models like BERT/GPT benefit from the pre-training stage on large-scale unsupervised data for better representation learning. Inspired by this, we explore how to simulate a pre-training stage for the compressed models. One method is to leverage large-scale unsupervised data of specific task for knowledge distillation. However it is usually hard to obtain large-scale task-specific unsupervised data for NLP tasks, such as NLI tasks from GLUE datasets. To address this challenge, a Q\&A  knowledge distillation task is proposed to pre-train the compressed student model on a large-scale Q\&A unlabeled data which are derived from a commercial search engine. To be more specific: 
\begin{itemize}
    \item \textit{Step 1:} For each question, top 10 relevant documents are returned by the commercial search engine to form <Question, Url> pairs, and passages are further extracted from these documents to form <Question, Passage> pairs. 
    \item \textit{Step 2:} Then we leverage several Q\&A teacher models (such as BERT\textsubscript{large} fine-tuned models) to score the above <Question, Passage> pairs.
    \item \textit{Step 3:} We use the <Question, Passage> corpus as well as their corresponding teacher models' output scores as the pseudo ground truth to pre-train the student model\footnote{The BERT student model is initialized by the bottom three layers of the BERT model. Therefore, it has captured a rough language model from large corpus.}.
\end{itemize}
With \textit{Step 1} and \textit{Step 2}, we could collect a large-scale auto labelled corpus (i.e. soft labels) for pre-training, which is several magnitudes larger than that of the human labeled training set. For \textit{Step 3}, we propose the novel multi-teacher knowledge distillation (i.e. m-o-1 approach) for pre-training. The distillation pre-trained student model\footnote{The distillation pre-trained model of stage 1 will be released soon.} with Q\&A task not only greatly boosts final Q\&A fine-tuned model but also other NLU tasks (like NLI tasks from GLUE), which are shown in experiment section later.


\subsubsection{Stage 2 - Task Specific Distillation Fine-tuning}
Through the large-scale distillation pre-training stage, our student model is able to learn decent feature representation capabilities for general NLU tasks (like Web Q\&A task, MNLI, SNLI, RTE tasks from GLUE). At the fine-tuning stage, the student model is firstly initialized with the pre-trained parameters in the above \textit{Stage 1}, and then all of the parameters are fine-tuned using labeled data from the downstream specific tasks. At this stage, we propose a novel multi-teacher knowledge distillation method (i.e. m-o-1 approach).

To be more specific, for each downstream task, we use both the golden label (i.e. ground-truth knowledge of $T_0$) on the task specific corpus and the soft labels of $T_1$-$T_N$ (i.e. pseudo ground-truth knowledge) on the same corpus to jointly fine-tune to get an enhanced student model. This is just like the learning process of human beings that we simultaneously gain knowledge from our teachers as well as the textbooks that our teachers have studied.

\subsection{TMKD Architecture}
TMKD is implemented from BERT~\cite{devlin2018bert}. Our model consists of three layers: Encoder layer utilizes the lexicon to embed both the question and passage into a low embedding space; Transformer layer maps the lexicon embedding to contextual embedding; Multi-header layer jointly learns from multiple teachers simultaneously during training, as well as generates final prediction output during inference. 

\subsubsection{Encoder Layer}
In Q\&A system, each question and passage are described by a set of words. We take the word pieces as the input just like BERT. $X=\{x^{(1)}, x^{(2)}, ..., x^{(|X|)}\}$ is to denote all the instances, and each instance has a $\left \langle Q, P \right \rangle$ pair. Let $Q={\{w_1, w_2, w_3, ..., w_m\}}$ be a question with $m$ word pieces, $P={\{w_1, w_2, w_3, ..., w_n\}}$ be a passage with $n$ word pieces, and $w_i$ is the bag-of-word representation of $i$-th word piece. $C = \{c_1, c_2, \dots, c_{|C|}\}$ represents the label set to indicate $\left \langle Q, P \right \rangle$'s relation. Each token representation is constructed by the sum of the corresponding token, segment and position embeddings. Let $\textbf{V}=\{\vec{v}_t \in \mathbb{R}^{D_v} | t = 1, \dots, M\}$ denote all the summed vectors in a $D_v$ dimension continuous space.

We concatenate the $\left \langle Q, P \right \rangle$ pair, and add $\left \langle CLS \right \rangle$ as the first token, then add $\left \langle SEP \right \rangle$ between Q and P. After that, we obtain the concatenation input $x_c = {\{w_1, w_2, w_3, \dots, w_{m+n+2}\}}$ of a given instance $x^{(i)}$. With the encoder layer, we map $x_c$ into continuous representations $H_e = {\{v_1, v_2, \dots, v_{m+n+2}\}}$. 

\subsubsection{Transformer Layer}
We also use the bidirectional transformer encoder to map the lexicon embedding $H_e$ into a sequence of continuous contextual embedding $H_s = {\{h_1, h_2, h_3, \dots, h_{m+n+2}\}}$. 

\subsubsection{Multi-header Layer}
In our proposed approach, firstly several teacher models are built with different hyper-parameters. Then, in order to let the student model to jointly learn from these teacher models, a multi-header layer is designed consisting of two parts, i.e. golden label header and soft label headers:

\paragraph{Golden Label Header} Given instance $x^{(i)}$, this header aims to learn the ground truth label. Following the BERT, we select $x^{(i)}$'s first token's transformer hidden state $h_1$ as the global representation of input. The probability that $x^{(i)}$ is labeled as class c is defined as follows:
\begin{equation}
    P(c|\left \langle Q, P \right \rangle) = softmax(W^{T}_g\cdot h_1)
\end{equation}
where $W^{T}_g$ is a learnable parameter matrix, $c \in C$ indicates the relation between $\left \langle Q, P \right \rangle$. The objective function of golden label header task is then defined as the cross-entropy:
\begin{equation}\label{eq:gold_loss}
    l_g = -\sum\limits_{c \in C}c \cdot log(P(c|\left \langle Q, P \right \rangle))
\end{equation}



\paragraph{Soft Label Headers} Take the $i$-th soft label as an example, $i in [1, |N|]$, N is the number of soft labels. For a given instance $x^{(i)}$, we also select the first token's hidden state $h_1$ as the global representation of input. The probability that $x^{(i)}$ is labeled as class $c$ is defined as follows:
\begin{equation}
    P_{s_{i}}(c|\left \langle Q, P \right \rangle) = softmax(W^{T}_{s_{i}}\cdot h_1)
\end{equation}
where $W^{T}_{s_i}$ is a learnable parameter matrix. We support $R_{s_i}(c|\left \langle Q, P \right \rangle) \\ \!=\!W^{T}_{s_i}\cdot h_1$ as the logits of $i$-th soft header before normalization.

For a instance $\left \langle Q, P \right \rangle$, teacher model can predict probability distributions to indicate that Q and P are relevant or not. Soft label headers aim to learn the teachers' knowledge through soft labels. The objective function of soft label headers is defined as mean squared error as follows:
\begin{table*}[t!]
    \small
    \centering
    \caption{\label{t:statistic}Statistics of experiment datasets (For DeepQA dataset, we have a test dataset, which is non-overlapping with the training set. For GLUE, please note that the results on development sets are reported, since GLUE does not distribute labels for the test sets).}

\begin{tabular}{c|ccc}
\hline
\textbf{Dataset}          & \textbf{\begin{tabular}[c]{@{}c@{}}Size of Samples\\ (Train/Test)\end{tabular}} & \textbf{\begin{tabular}[c]{@{}c@{}}Average Question Length\\ (Words)\end{tabular}} & \textbf{\begin{tabular}[c]{@{}c@{}}Average Answer Length\\ (Words)\end{tabular}} \\ \hline
\textbf{DeepQA}           & 1M/10K                                                                          & 5.86                                                                               & 43.74                                                                            \\
\textbf{CommQA-Unlabeled} & 4M(base) 40M(large) 0.1B(extreme)                 & 6.31                                                                               & 42.70                                                                            \\
\textbf{CommQA-Labeled}   & 12M/2.49K                                                                       & 5.81                                                                                & 45.70                                                                              \\ 
\textbf{MNLI}             & 392.70K/19.64K                                                                  & 20.52                                                                              & 10.90                                                                            \\
\textbf{SNLI}             & 549.36K/9.84K                                                                   & 13.80                                                                              & 10.90                                                                            \\
\textbf{QNLI}             & 108.43K/5.73K                                                                   & 9.93                                                                               & 28.07                                                                            \\
\textbf{RTE}              & 2.49K/0.27K                                                                     & 45.30                                                                              & 9.77                                                                             \\ \hline
\end{tabular}
\vspace{-8pt}
\end{table*}


\begin{equation}\label{eq:score_loss}
    \begin{aligned}
        &l_{s_i} = \frac{1}{|C|}\sum_{c \in C} (R_{s_i}(c|\left \langle Q, P \right \rangle) - R_{t_i}(c|\left \langle Q, P \right \rangle))^2 \\
        &l_s = \frac{1}{N}\sum_{i=1}^{N}l_{s_i} \\            
    \end{aligned}
\end{equation}
where $R_{t_i}(c|\left \langle Q, P \right \rangle)$ represents the $i$-th soft label teacher's logits before normalization and $N$ is the number of soft label headers. 

\subsection{Training and Prediction}

\label{sec:train_and_prediction}
In order to learn parameters of \textbf{TMKD} model, our proposed TMKD model has a two-stage training strategy. At the first stage, we use the Equation~(\ref{eq:score_loss}) to learn the generalized natural language inference capability from the unlabeled data with soft labels. At the second stage, we combine Equation~(\ref{eq:gold_loss}) and Equation~(\ref{eq:score_loss}) to learn the task-specific knowledge from the labeled data with golden labels and soft labels, then obtain our final learning objective function as follows:
\begin{equation}\label{eq:multi}
    l = (1-\alpha)l_g + \alpha l_s
\end{equation}
where $\alpha$ is a loss weighted ratio, $l_{s_{i}}$ is the loss of $i$-th soft header. In the inference stage, we use an aggregation operation to calculate the final result as follows:
\begin{equation}
    \begin{aligned}
        O(c|\left \langle Q, P \right \rangle) = & \frac{1}{N+1}(P(c|\left \langle Q, P \right \rangle) + \\ & \sum_{i=1}^N P_{s_i}(c|\left \langle Q, P \right \rangle))
    \end{aligned}
\end{equation}
where $P_{s_i}$ is the $i$-th student header's output and $N$ denotes the number of soft label headers.



\section{Experiment}
In this section, we conduct empirical experiments to verify the effectiveness of our proposed TMKD on model compression. We first introduce the experimental settings, then compare our model to the baseline methods to demonstrate its effectiveness.

\subsection{Dataset}
\label{sec:dataset}
We conduct experiments on several datasets as follows. 

\begin{itemize}
    \item \textbf{DeepQA}: An English Q\&A task dataset from one commercial Q\&A system, with 1 million labeled cases. Each case consists of three parts, i.e. question, passage, and binary label (i.e. 0 or 1) by crowd sourcing judges indicating whether the question can be answered by the passage. The following briefly describes how the data is collected. Firstly, for each question, top 10 relevant documents returned by the search engine are selected to form <Question, Url> pairs; Then passages are further extracted from these documents to form <Question, Url, Passage> triples; These <Query, Passage> pairs are sampled and sent to crowd sourcing judges. Specifically, each <Query, Passage> pair is required to get judged by three judges. Those cases with more than 2/3 positive labels will get positive labels, otherwise negative.  
    \item \textbf{CommQA-Unlabeled} A large-scale unlabeled Q\&A data coming from a commercial search engine. The collection method of <Query, Passage> pairs is same as DeepQA, and the difference is that the question type and domain of this dataset is more diverse than DeepQA. We sampled 4 million (named base dataset) and 40 million (named large dataset) as the pre-training data. Besides, in our commercial scenario, we have one extremely large Q\&A unlabeled dataset (0.1 billion) cooked by the same data collection approach.
    \item \textbf{CommQA-Labeled} A large-scale commercial Q\&A training data, which is sampled from CommQA-Unlabeled, and labeled by crowd sourcing judges.
    \item \textbf{GLUE} \cite{WangSMHLB18}: A collection of datasets for evaluating NLU systems, including nine language understanding tasks. Among them, we choose textual entailment tasks (MNLI, SNLI, QNLI, RTE), which are similar to Q\&A task. For MNLI and QNLI, given two sentences (premise and hypothesis), the task is to predict whether the premise entails the hypothesis (entailment), contradicts (contradiction), or neither (neutral). While for SNLI and RTE, the relationship does not contain neutral type.
\end{itemize}

\begin{table*}[t!]
\centering
\small
\caption{\small\label{t:main}Model comparison between our methods and baseline methods. ACC denotes accuracy (all ACC metrics in the table are percentage numbers with \% omitted). Specially for MNLI, we average the results of matched and mismatched validation set. }

\begin{tabular}{ll|rrrrr|rr}
\hline
\multicolumn{2}{c|}{\multirow{2}{*}{\textbf{Model}}}                                                                  			& \multicolumn{5}{c|}{\textbf{Performance (ACC)}}                                                                        & \multirow{2}{*}{\textbf{\begin{tabular}[c]{@{}c@{}}Inference \\ Speed(QPS)\end{tabular}}} & \multirow{2}{*}{\textbf{\begin{tabular}[c]{@{}c@{}}Parameters\\ (M)\end{tabular}}} \\
\multicolumn{2}{c|}{}                                                                                                 			& \textbf{DeepQA}      & \textbf{MNLI}        & \textbf{SNLI}        & \textbf{QNLI}        & \textbf{RTE}         &                                                                                           &                                                                                    \\ \hline
\multirow{3}{*}{\textbf{Original Model}}                                                    & \textbf{BERT-3} 	& 75.78                & 70.77                & 77.75                & 78.51                & 57.42                & 207                                                                                       & 50.44                                                                              \\ 
         & \textbf{BERT\textsubscript{large}}     	& 81.47                & 79.10                & 80.90                & 90.30                & 68.23                & 16                                                                                        & 333.58                                                                             \\ 
		   								       & \textbf{BERT\textsubscript{large} ensemble}    & 81.66                & 79.57                & 81.39                & 90.91                & 70.75                & 16/3                                                                                        & 333.58*3                                                                             \\ \hline
\multirow{6}{*}{\textbf{\begin{tabular}[c]{@{}c@{}}Traditional Distillation\\ Model\end{tabular}}} & \textbf{Bi-LSTM (1-o-1)}      			& 71.69                & 59.39                & 69.59                & 69.12                & 56.31                & 207                                                                                       & 50.44                                                                              \\
                                                                                       & \textbf{Bi-LSTM (1\textsubscript{avg}-o-1)}    & 71.93           	       & 59.60                   & 70.04                   & 69.53                & 57.35                & 207                                                                                       & 50.44  \\
                                                                                       & \textbf{Bi-LSTM (m-o-m)}      			& 72.04                & 61.71                & 72.89                & 69.89                & 58.12                & 207/3                                                                                       & 50.44*3                                                                            \\
                                                                                       & \textbf{BERT-3 (1-o-1)}       			& 77.35                & 71.07                & 78.62                & 77.65                & 55.23                & 217                                                                                       & 45.69                                                                              \\
                                                                                       & \textbf{BERT-3 (1\textsubscript{avg}-o-1)}     & 77.63                 & 70.63			& 78.64			& 78.20                & 58.12                & 217                                                                                       & 45.69                                                                              \\
                                                                                       & \textbf{BERT-3 (m-o-m)}       			& 77.44                & 71.28                & 78.71                & 77.90                & 57.40                & 217/3                                                                                     & 45.69*3                                                                            \\ \hline
\multirow{4}{*}{\textbf{Our Distillation Model}}                                                    & \textbf{Bi-LSTM (TMKD\textsubscript{base})} 	& 74.73                & 61.68                & 71.71                & 69.99                & 62.74                & 207                                                                                       & 50.45                                                                              \\
                                                                                       & \textbf{$^*$TMKD\textsubscript{base}}         	& 79.93                & 71.29                & 78.35                & 83.53                & 66.64                & 217                                                                                       & 45.70                                                                              \\
                                                                                       & \textbf{$^*$TMKD\textsubscript{large}}         	& {\ul \textbf{80.43}} & {\ul \textbf{73.93}} & {\ul \textbf{79.48}} & {\ul \textbf{86.44}} & {\ul \textbf{67.50}} & {\ul \textbf{217}}                                                                        & {\ul \textbf{45.70}}                                                               \\ \hline
\end{tabular}

\footnotesize{$^*$ These two models are BERT-3 based models.}\\

\end{table*}

\subsection{Evaluation Metrics}
We use the following metrics to evaluate model performance:
\begin{itemize}
    \item \textbf{Accuracy (ACC)}: Number of correct predictions divided by the total number of samples.
    \item \textbf{Queries Per Second (QPS)}: Average number of cases processed per second. We use this metric to evaluate the model inference speed.
\end{itemize}

\subsection{Baselines}
We compare our model with several strong baselines to verify the effectiveness of our approach.
\begin{itemize}
    \item \textbf{BERT-3}: a student model without any knowledge distillation but instead trained as a small version of BERT/GPT, which initialized by the bottom 3-layer weight of BERT.
    \item \textbf{BERT\textsubscript{large}~\cite{devlin2018bert}}: We use the BERT\textsubscript{large} fine-tuning model (24-layer transformer blocks, 1024 hidden size, and 16 heads) as another strong baseline.
    \item \textbf{BERT\textsubscript{large} Ensemble}: We use BERT\textsubscript{large} fine-tuning model ensemble as another strong baseline (the output probability distribution decided by the average probability distributions of all models).
    \item \textbf{Single Student Model~(1-o-1 and 1\textsubscript{avg}-o-1)~\cite{hinton2015distilling}}:  Student model learns from one single teacher model using knowledge distillation. For teacher model selection, we have two strategies. Firstly, we pick the best model selected from \textbf{Original BERT} teacher models to distill one single model (called 1-o-1). Secondly, we pick the average score of teacher models as another special teacher to distill one single student (called 1\textsubscript{avg}-o-1). We implement this method under two architectures: BERT-3 model and Bi-LSTM model. In the following sections, where we do not clarify the basic model is BERT-3 model.
    \item \textbf{Student Model Ensemble (m-o-m)}: For each teacher model, 1-o-1 is used to train a single student model. Based on this method, 3 separate student models are trained based on 3 different teacher models. Finally an ensemble aggregation is used by simply averaging the output scores to form the final results. We also implement it under BERT-3 base model and Bi-LSTM model. 
\end{itemize}


\subsection{Parameter Settings}
For all BERT based models, we implement on top of the PyTorch implementation of BERT\footnote{github.com/huggingface/pytorch-pretrained-BERT.}. All teacher models are trained using BERT\textsubscript{large} with batch size of 128 for 10 epochs, and max sequence length as 150. On each dataset, we train three different teacher models with different learning rates in $\{2, 3, 5\} \times 10^{-5}$. For BERT-3 student model, we optimize the student model using a learning rate of $1 \times 10^{-4}$, and all BERT-based models are initialized using pre-trained BERT model weights.

For all Bi-LSTM based models, we set the LSTM hidden units as 256, LSTM layer count as 2, and word embedding dimension as 300. Top 15 thousands of words are selected as vocabulary and 300 dimension Glove is used for embedding weight initialization. Words not in Glove vocabulary are randomly initialized with normal distribution. The parameters are optimized using Adam optimizer with learning rate as $1 \times 10^{-3}$. 

Those teacher models used for TMKD and m-o-m training are identical for fair comparison. The only difference between TMKD\textsubscript{base} and TMKD\textsubscript{large} is the training data in the distillation pre-training stage. To be more specific, TMKD\textsubscript{base} leverages CommQA-Unlabeled base corpus for pre-training while TMKD\textsubscript{large} is pre-trained using CommQA-Unlabeled large corpus. 

\subsection{Comparison Against Baselines}
In this section, we conduct experiments to compare TMKD with baselines in terms of three dimensions, i.e. inference speed, parameter size and performance on task specific test set. From the results shown in Table~\ref{t:main}, it is intuitive to have the following observations:
\begin{itemize}
    \item It is not surprising that original BERT teacher model shows the best performance due to its sheer amount of parameters (340M), but inference speed is super slow and memory consumption is huge for production usage. 
    \item 1-o-1 and 1\textsubscript{avg}-o-1 (BERT-3 and Bi-LSTM) obtain pretty good results regarding inference speed and memory capacity. However there are still some gaps compared to the original BERT model in terms of ACC metric. 
    \item m-o-m performs better than 1-o-1. However, the inference speed and memory consumption increase in proportion to the number of student models used for ensemble. 
    \item Compared with 1-o-1, 1\textsubscript{avg}-o-1 and m-o-m, TMKD achieves optimum in all three dimensions. In terms of memory, TMKD only needs small amount of additional memory consumption since the majority of parameters are shared across different distillation tasks compared with the 1-o-1. In addition, TMKD performs significant better than BERT-3, which further proves the effective of our model.
\end{itemize}

To conclude, TMKD performs better in three dimensions than several strong baseline compressed models with knowledge distillation (i.e. 1-o-1, 1\textsubscript{avg}-o-1, m-o-m) on all the evaluation datasets, and also further decreases performance gap with the original BERT model, which verifies the effectiveness of TMKD. 



 
\section{Ablation Studies}
TMKD consists of multiple teacher distillation pre-training stage and distillation fine-tuning stage. In this section, we further conduct several experiments to analyze the contribution of each factor in TMKD, in order to obtain a better understanding of the proposed approach.


\begin{figure*}[t!]
    \centering
    \includegraphics[scale=0.37, viewport=15 180 1400 545, clip=true]{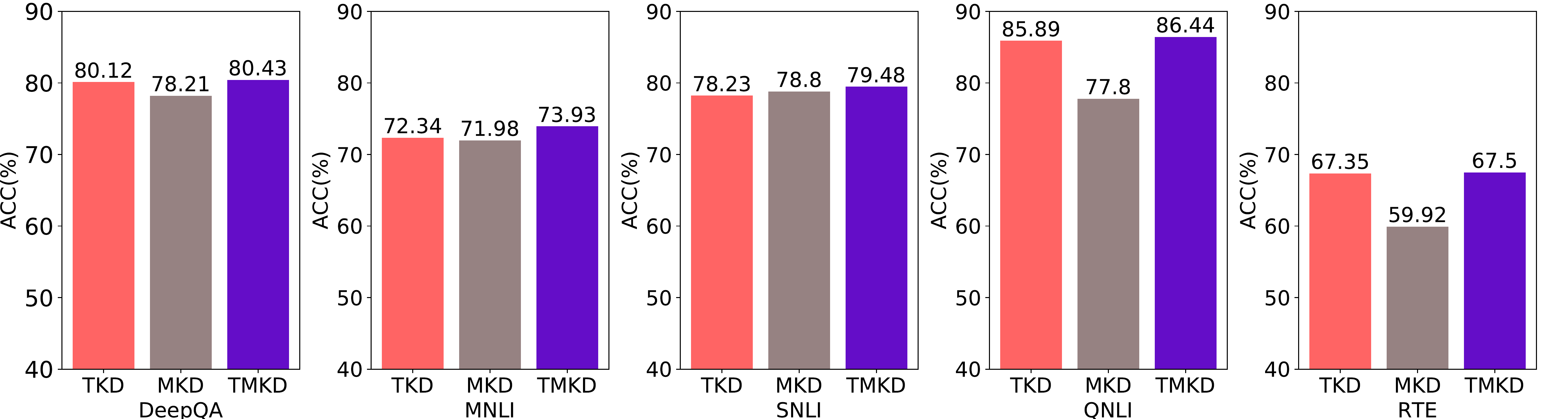}
    \caption{\label{fig:sub_model} Performance comparison of TKD, MKD and TMKD on different datasets}
\vspace{-8pt}
\end{figure*}

\subsection{Impact of Different Training Stages}
\label{sec:stages}

\subsubsection{Impact of Distillation Pre-training Stage.}
One advantage of TMKD is to introduce a multi-teacher distillation task for student model pre-training to boost model performance. We analyze the impact of pre-training stage by evaluating two new models:

\textbf{TKD}: A 3-layer BERT\textsubscript{base} model which is firstly trained using 1-o-1 distillation pre-training on CommQA-Unlabeled large-scale dataset (i.e. 40M <Question, Passage> pairs), then fine-tuned on task specific corpus with golden label and single soft label (i.e. by only one teacher) of each task. 

\textbf{KD (1-o-1)}: Another 3-layer BERT\textsubscript{base} model which is fine-tuned on task specific corpus with golden label and single soft label of each task but without distillation pre-training stage. 

From the results in Table~\ref{t:stage1_com}, we have the following observations: (1) On DeepQA dataset, TKD shows significant gains by leveraging large-scale unsupervised Q\&A pairs for distillation pre-training. (2) Although Q\&A task is different with GLUE tasks, the student model of GLUE tasks still benefit a lot from the distillation pre-training stage leveraging Q\&A task. This proves the effect of the distillation pre-training stage leveraging Q\&A large corpus. 

\begin{table}[htbp]
    \small
    \caption{\label{t:stage1_com} Comparison between KD and TKD}
    \centering
    \begin{tabular}{@{}lrrrrr@{}}
    \toprule
    \textbf{Model} & \multicolumn{5}{c}{\textbf{Performance (ACC)}}                                       \\
                   & \textbf{DeepQA} & \textbf{MNLI} & \textbf{SNLI} & \textbf{QNLI} & \textbf{RTE} \\ \midrule
    \textbf{KD (1-o-1)}                     & 77.35             & 71.07             & 78.62             & 77.65             & 55.23          \\
    \textbf{TKD}  & {\ul \textbf{80.12}}      & {\ul \textbf{72.34}}             & {\ul \textbf{78.23}}           & {\ul \textbf{85.89}}         & {\ul \textbf{67.35}}   \\ \bottomrule
    \end{tabular}
    \vspace{-10pt}
\end{table}

\subsubsection{Impact of Multi-teacher Distillation vs Single-teacher Distillation.} Another advantage of TMKD is designing a unified framework to jointly learn from multiple teachers. We analyze the impact of multi-teacher versus single-teacher knowledge distillation by the following three models:
 
 \textbf{MKD}: A 3-layer BERT\textsubscript{base} model trained by Multi-teacher distillation (m-o-1) without pre-training stage.  
 
\textbf{KD (1\textsubscript{avg}-o-1)}: A 3-layer BERT\textsubscript{base} model trained by Single-teacher distillation (1\textsubscript{avg}-o-1) without pre-training stage, which is to learn from the average score of teacher models.
\begin{table}[htbp]
    \small
    \caption{\label{t:stage2_com} Comparison Between KD (1\textsubscript{avg}-o-1) and MKD}

    \centering
    \begin{tabular}{@{}lrrrrr@{}}
    \toprule
    \multirow{1}{*}{\textbf{Model}} & \multicolumn{5}{c}{\textbf{Performance (ACC)}}                                       \\
                                    & \textbf{DeepQA}   & \textbf{MNLI}     & \textbf{SNLI}     & \textbf{QNLI}     & \textbf{RTE} \\ \midrule
    \textbf{KD (1\textsubscript{avg}-o-1)}  & 77.63             & 70.63             & 78.64             & {\ul \textbf{78.20}}             & 58.12 \\
    \textbf{MKD}                    & {\ul \textbf{78.21}}             & {\ul \textbf{71.98}}           & {\ul \textbf{78.80}}           & 77.80           & {\ul \textbf{59.92}}          \\ \bottomrule
    \end{tabular}
        \vspace{-8pt}
\end{table}

From Table~\ref{t:stage2_com}, MKD outperforms KD (1\textsubscript{avg}-o-1) on the majority of tasks, which demonstrates that multi-teacher distillation approach (m-o-1) is able to help student model learn more generalized knowledge by fusing knowledge from different teachers.

\subsubsection{Dual-Impact of Two Stages.}
Finally, TKD, MKD and TMKD are compared altogether. From Figure~\ref{fig:sub_model}, TMKD significantly outperforms TKD and MKD in all datasets, which verifies the complementary impact of the two stages (distillation pre-training \& m-o-1 fine-tuning) for the best results.

\subsubsection{Extensive Experiments: Multi-teacher Ensemble or Multi-teacher Distillation?}

TMKD leverage multi-teacher distillation in both pre-training and task specific fine-tuning stages. This multi-teacher mechanism actually introduces multi-source information from different teachers. A common approach to introduce multi-source information is \emph{ensemble} (e.g. average score of the prediction outputs from multiple models). Compared with the common multi-teacher ensemble approach, are there extra benefits from multi-teacher distillation? We conduct further experiments to explore this question.

For clear comparisons, we apply some degradation operations to TMKD. We remove the multi-teacher distillation mechanism from TMKD, and then use ensemble teacher score (the average score of soft labels by multiple teachers) and single teacher score (from the best teacher) to train two new models with a two-stage setting respectively, which are denoted as TKD\textsubscript{base} (1\textsubscript{avg}-o-1) and TKD\textsubscript{base} (1-o-1). Experiments using both BERT-3 and Bi-LSTM as the student model architecture are conducted, as shown in Table~\ref{t:TKD_com}.

\begin{table}[htbp]
\small
\caption{\label{t:TKD_com} Comparison between TKD and TMKD}
\centering
\renewcommand\tabcolsep{2.9pt}
\begin{tabular}{lrrrrr}
\hline
\textbf{Model}                              		    & \multicolumn{5}{c}{\textbf{Dataset}}                                           \\
\textbf{}                                   		    & \textbf{DeepQA}   & \textbf{MNLI}     & \textbf{SNLI}     & \textbf{QNLI}     & \textbf{RTE} \\ \hline
\textbf{Bi-LSTM (TKD\textsubscript{base} (1-o-1))}    		    & 74.26           	& 61.43             & 71.54             & 69.2              & 59.56        \\
\textbf{Bi-LSTM (TKD\textsubscript{base} (1\textsubscript{avg}-o-1))} 		    & 74.38           	& 61.55             & 71.7              & 69.08             & 61.01        \\
\textbf{Bi-LSTM (TMKD\textsubscript{base})}              & {\ul\textbf{74.73}}           	& {\ul\textbf{61.68}}             & {\ul\textbf{71.71}}             & {\ul\textbf{69.99}}             & {\ul\textbf{62.74}}        \\ \hline
\textbf{$^*$TKD\textsubscript{base} (1-o-1)}             	 	    & 79.5     		    & 71.07             & 77.66             & 82.79             & 63.89        \\
\textbf{$^*$TKD\textsubscript{base} (1\textsubscript{avg}-o-1)}           		    & 79.73           	& 71.21             & 77.70              & 83.40              & {\ul\textbf{67.10}}         \\
\textbf{$^*$TMKD\textsubscript{base}}                    	& {\ul\textbf{79.93}}            	& {\ul\textbf{71.29}}             & {\ul\textbf{78.35}}             & {\ul\textbf{83.53}}             & 66.64        \\ \hline
\end{tabular}
\footnotesize{$^*$ These three models are BERT-3 based models.}\\

\end{table}

From the results, we have the following observations: (1) For both BERT-3 ad Bi-LSTM based models, the TKD\textsubscript{base}(1\textsubscript{avg}-o-1) performs better than TKD\textsubscript{base}(1-o-1). This demonstrates that ensemble of teacher models is able to provide more robust knowledge than single teacher model when distill the student model. (2) Compared with TKD\textsubscript{base}(1-o-1) and TKD\textsubscript{base}(1\textsubscript{avg}-o-1), TMKD\textsubscript{base} obtains the best performance no matter using Bi-LSTM or BERT-3. It because that the multi-source information was diluted by the average score. TMKD introduces the differences when training, the multi-source information can be adpative at the training stage.

\subsection{Impact of Training Data Size}
To further evaluate the potential of TMKD, we conduct extensive experiments on CommQA-Unlabeled extremely large-scale corpus data (0.1 billion unlabeled Q\&A pairs) and CommQA-Labeled (12M labeled Q\&A labeled pairs). Four separate teacher models ($T_1$-$T_4$) are trained with batch size of 128 and learning rate with $\{2,3,4,5\}*e^{-5}$. Max sequence length is set as 200, and number of epochs as 4. The settings of KD, MKD, and TMKD keep the same as Section~\ref{sec:stages}. The results are shown in Table~\ref{t:online_exp}. Interestingly, on this extremely large Q\&A dataset, TMKD even exceeds the performance of its teacher model (ACC: 79.22 vs 77.00), which further verifies the effectiveness of our approach. 

\begin{table}[htbp]
    \small
    \centering
    \caption{\label{t:online_exp} Extremely large Q\&A dataset results. }
\vspace{-6pt}
\begin{tabular}{ccccc}
\hline
& \multicolumn{4}{c}{\textbf{Performance (ACC)}}                          \\ 
                & \textbf{BERT\textsubscript{large}} & \textbf{KD} & \textbf{MKD} & \textbf{TMKD} \\\hline
& 77.00               & 73.22        & 77.32        & {\ul \textbf{79.22}}         \\ \hline
\end{tabular}
\vspace{-12pt}
\end{table}

\subsection{Impact of Transformer Layer Count}
In this section, we discuss the impact of transformer layer count $n$ for TMKD\footnote{In order to save experimental costs, we choose TMKD\textsubscript{base} for experimentation.} with  $n \in \{1, 3, 5\}$. As observed from Table~\ref{t:different_layer}: (1) With $n$ increasing, ACC increases as well but inference speed decreases, which aligns with our intuition.
(2) With $n$ increasing, the performance gain between two consecutive trials decreases. That say, when $n$ increases from $1$ to $3$, the ACC gains of the 5 datasets are ($3.87$, $9.90$, $7.46$, $11.44$, $11.19$) which are very big jump; while $n$ increases from $3$ to $5$, gains decrease to ($1.08$, $1.63$, $0.53$, $2.89$, $0.37$), without decent add-on value compared with the significantly decreased QPS. 

\begin{table}[htbp]
    \small
    \centering
    \caption{\label{t:different_layer} Compare different number of transformer layer.}
\vspace{-6pt}
\begin{tabular}{l|l|rrr}
\hline
\textbf{Dataset} & \textbf{Metrics} & \multicolumn{3}{c}{\textbf{Layer Number}} \\
\textbf{}        & \textbf{}        & \textbf{1}   & \textbf{3}   & \textbf{5}  \\ \hline
\textbf{DeepQA}  & ACC              & 74.59          & 78.46          & 79.54         \\
\textbf{MNLI}    & ACC              & 61.23          & 71.13          & 72.76         \\
\textbf{SNLI}    & ACC              & 70.21          & 77.67          & 78.20         \\
\textbf{QNLI}    & ACC              & 70.60          & 82.04          & 84.94         \\
\textbf{RTE}     & ACC              & 54.51          & 65.70          & 66.07         \\ \hline
                 & QPS              & 511          & 217          & 141         \\ \hline
\end{tabular}
\vspace{-6pt}
\end{table}

Based on the above results, we set $n$ as $3$ since it has the highest performance/QPS ratio for web Question Answering System. In real production scenarios, we need to balance between performance and latency. 
\subsection{Impact of Loss Weighted Ratio}

\begin{figure}[t!]
    \centering
    \includegraphics[scale=0.42, viewport=10 7 455 330, clip=true]{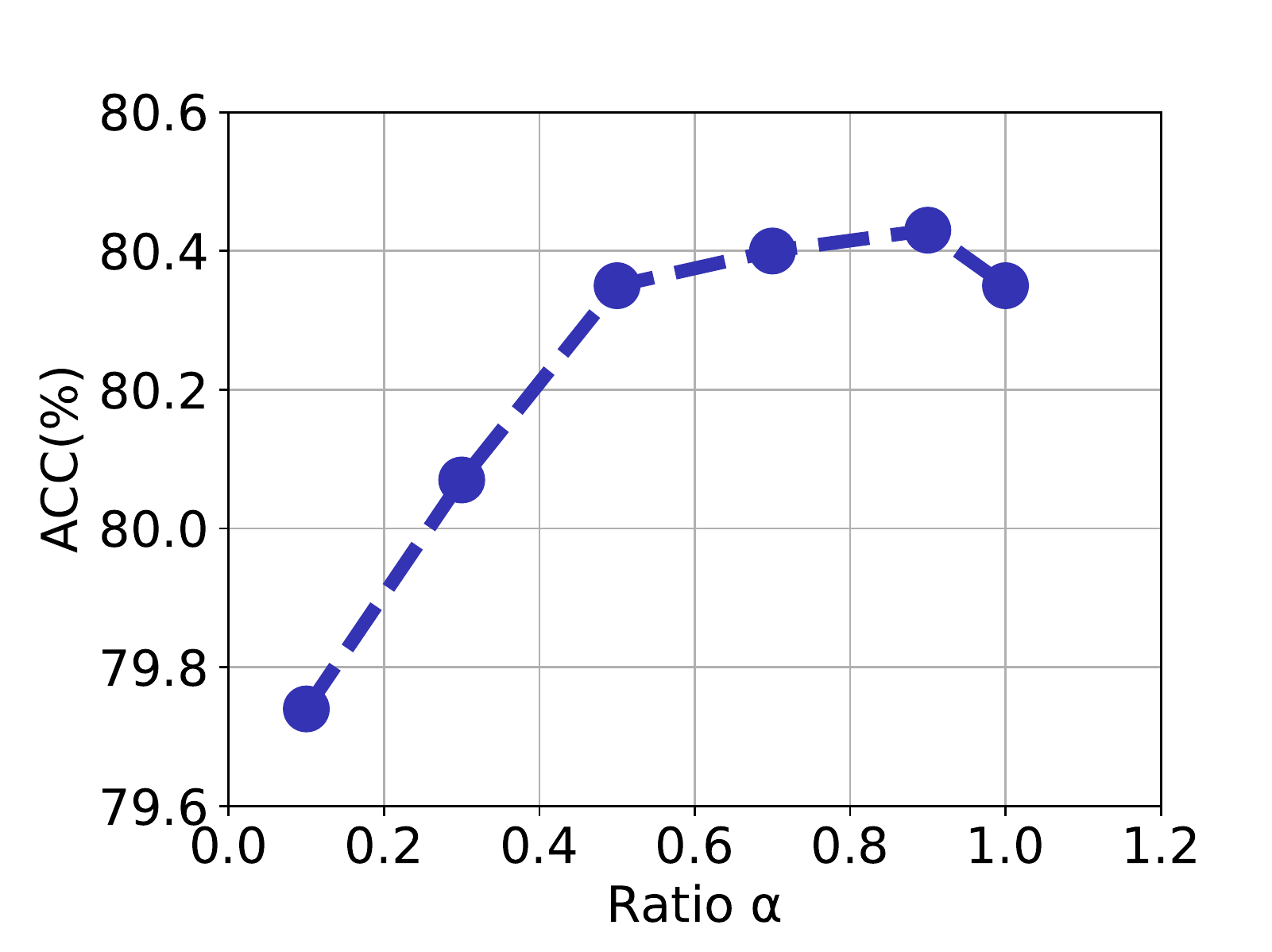}
    \vspace{-4pt}
    \caption{\label{fig:different_ratio} The impact of different loss weighted ratio.}
    \vspace{-15pt}
\end{figure}

We also conducts several experiments to analyze the impact of the loss weighted ratio $\alpha$ defined in Section~\ref{sec:train_and_prediction}, where $\alpha \in \{0.1, 0.3, 0.5, \\ 0.7, 0.9, 1.0\}$. Specially, when set the ratio as $1.0$, we only use the soft label headers to calculate the final output result. The results of TMKD against different $\alpha$ values are shown in Figure~\ref{fig:different_ratio}. We can observe: 
(1) The larger value the ratio is, the better performance is obtained (except when $\alpha$ is $1.0$). (2) Without the golden label supervision (i.e. $\alpha$ is $1.0$), the performance decreases. The intuition is just like the knowledge learning process of human beings. We learn knowledge not only from teachers but also through reading books which can provide us a comprehensive way to master knowledge with less bias.
\section{Conclusion and Future Work}


In this paper, we propose a novel Two-stage Multi-teacher Knowledge Distillation (\textbf{TMKD}) approach for model compression. Firstly a Q\&A multi-teacher distillation task is proposed for student model pre-training, then a multi-teacher paradigm is designed to jointly learn from multiple teacher models (m-o-1) for more generalized knowledge distillation on downstream specific tasks. Experiment results show that our proposed method outperforms the baseline state-of-art methods by great margin and even achieves comparable results with the original teacher models, along with significant speedup of model inference. The compressed Q\&A model with TMKD has already been applied to real commercial scenarios which brings significant gains.

In the future, we will extend our methods to more NLU tasks, such as sequence labelling, machine reading comprehension, etc. On the other hand, we will explore how to select teacher models more effectively for better student model distillation.

\bibliographystyle{ACM-Reference-Format}
\bibliography{paper_reference}

\end{document}